# Approximating the Backbone in the Weighted Maximum Satisfiability Problem[*]

JIANG He[12+], XUAN Ji-Feng[1], HU Yan[1]

[1](School of Software, Dalian University of Technology, Dalian 116621, China)

[2](The State Key Laboratory of Computer Science, Institute of Software, CAS, Beijing 100190, China)

+ Corresponding author: Phn: +86-411-81381830, Fax: +86-411-87571567, E-mail: jianghe@dlut.edu.cn,

**Abstract.** The weighted Maximum Satisfiability problem (weighted MAX-SAT) is a NP-hard problem with numerous applications arising in artificial intelligence. As an efficient tool for heuristic design, the backbone has been applied to heuristics design for many NP-hard problems. In this paper, we investigated the computational complexity for retrieving the backbone in weighted MAX-SAT and developed a new algorithm for solving this problem. We showed that it is intractable to retrieve the full backbone under the assumption that $P \neq NP$. Moreover, it is intractable to retrieve a fixed fraction of the backbone as well. And then we presented a backbone guided local search (BGLS) with Walksat operator for weighted MAX-SAT. BGLS consists of two phases: the first phase samples the backbone information from local optima and the backbone phase conducts local search under the guideline of backbone. Extensive experimental results on the benchmark showed that BGLS outperforms the existing heuristics in both solution quality and runtime.

## 1. Introduction

The Satisfiability problem (SAT) is a famous NP-Complete problem [1], which consists of an assignment of Boolean variables (true or false) and some clauses formed of these variables. A clause is a disjunction of some Boolean literals and can be true if and only if any of them is true. A SAT instance is satisfied if and only if all the clauses are simultaneously true. As a generalization of SAT, the Maximum Satisfiability problem (MAX-SAT) aims to maximize the number of satisfied clauses. When every clause is associated with some weight, the MAX-SAT turns to the weighted Maximum Satisfiability problem (weighted MAX-SAT) with numerous applications arising in artificial intelligence, such as scheduling, data mining, pattern recognition, and automatic reasoning [2]. According to the computational complexity theory, there's no polynomial time algorithm for solving weighted MAX-SAT unless $P = NP$ [1]. Hence, many heuristic algorithms capable of finding near optimal solutions in reasonable time have been proposed for weighted MAX-SAT, including GSAT [4], GLS [5], Taboo Scatter Search [6], ACO [7], and GRASP/GRASP with path-relinking[8-9].

As an efficient tool for heuristic design, the backbone has attracted great attention from the society of artificial intelligence in recent years. The backbone is defined as the common parts of all optimal solutions for an instance. Since it's usually intractable to obtain the exact backbone, many approximate backbone guided heuristics have been developed for NP-Complete problems, including 3-SAT [10], TSP [11], QAP [12], et al. For SAT or MAX-SAT, some character and algorithms of backbone are in research [10, 13, 14]. And a BGWalksat (backbone guided local

[*] Supported by the National Research Foundation for the Doctoral Program of Higher Education of China under Grant No. 20070141020; the Natural Science Foundation of Liaoning Province of China under Grant No.20051082; the Gifted Young Foundation of Dalian University of Technology

search algorithm with Walksat operator) has been developed for MAX-SAT and achieves better performance than existing heuristics [14]. However, as to our knowledge, there is no theoretical result or backbone guided heuristics reported in the literature for weighted MAX-SAT.

In this paper, we investigated the computational complexity of the backbone for weighted MAX-SAT and developed a backbone guided local search (BGLS) algorithm. Firstly, we proved that there's no polynomial time algorithm to retrieve the full backbone under the assumption that $P \neq NP$. In the proof, we mapped any weighted MAX-SAT instance to a biased weighted MAX-SAT instance with a unique optimal solution by slightly perturbing, which is also optimal to the original instance. Based on this proof, we indicated that it's also intractable to retrieve a fixed fraction of the backbone, by reducing any weighted MAX-SAT instance to a series of weighted MAX-SAT instances with smaller scale. Secondly, we developed a backbone guided local search algorithm with pseudo-backbone frequencies. It consists of two phases: the sampling phase records local optima to compute pseudo-backbone frequencies and the backbone phase uses such information to guide the following local search. Experimental results demonstrated that BGLS obtained better solutions than GRASP/ GRASP with path-relinking in far less time.

This paper is organized as follows. In Section 2, we gave out the definitions to weighted MAX-SAT, the backbone and the biased weighted MAX-SAT. In Section 3, we gave the proof for the computational complexity of obtaining the backbone in weighted MAX-SAT. In Section 4, we described the backbone guided local search. The experimental results were presented in Section 5. Finally, we concluded the paper in Section 6.

## 2. Preliminaries

In this section, we shall give some definitions and related properties.

**Definition 1.** Given a set of Boolean variables $V = \{x_1, x_2, ..., x_n\}$, a propositional formula $\phi$ is a conjunction on a set of $m$ clauses $C = \{C_1, C_2, ..., C_m\}$. Each clause $C_i$ is a disjunction of $|C_i|$ literals, where each literal $l_{ij}$ is either a variable or its negation. A clause is satisfied if at least one of its literals evaluates to true, and the propositional formula $\phi$ is said to be satisfied if all of its clauses are satisfied. The *satisfiability* problem (SAT) aims to find an assignment of values to the variables such that a given propositional formula $\phi$ is satisfied. Formally, a SAT instance on the set of Boolean variables $V$ and the set of clauses $C$ can be denoted as $SAT(V, C)$.

**Definition 2.** Given a set of Boolean variables $V = \{x_1, x_2, ..., x_n\}$, a propositional weighted formula $\tilde{\phi}$ is a conjunction on a set of $m$ weighted clauses $\tilde{C} = \{\tilde{C}_1, \tilde{C}_2, ..., \tilde{C}_m\}$. Each weighted clause is associated with a positive weight $w(\tilde{C}_i)$. The *weighted maximum satisfiability* problem (weighted MAX-SAT) consists of finding an assignment of values to the variables such that the sum of the weights of the satisfied clauses is maximized. Similar to the SAT, a weighted MAX-SAT instance can be denoted as $wMAX - SAT(V, \tilde{C})$ and a *solution* (assignment) can be denoted as $s = \{x_i | x_i \text{ is assigned true}\} \bigcup \{\neg x_i | x_i \text{ is assigned false}\}$ with cost function $w(V, \tilde{C}, s) = \sum_{\tilde{C}_i \in \tilde{C}} w(\tilde{C}_i)[\tilde{C}_i \text{ is satified by s}]$, where $[\bullet] = 1$ (0) for $\bullet$ being true (false).

Without loss of generality, we shall assume in the following part of this paper, that the weight $w(\tilde{C}_i)$ is a positive integer for each clause $\tilde{C}_i$.

**Definition 3.** Given a weighted MAX-SAT instance $wMAX - SAT(V, \tilde{C})$, let $\Pi^* = \{s_1^*, s_2^*, ..., s_q^*\}$ be the set of all optimal solutions to it, where $q = |\Pi^*|$ represents the number of optimal solutions. *Backbone* in weighted MAX-SAT instance $wMAX - SAT(V, \tilde{C})$ is defined as $bone(V,\tilde{C}) = s_1^* \cap s_2^* \cap \cdots \cap s_q^*$.

The backbone $bone(V,\tilde{C})$ is essential for heuristic algorithms design, since a heuristic cannot obtain an optimal solution to a weighted MAX-SAT instance unless it retrieves the full backbone. In contrast, if the backbone $bone(V,\tilde{C})$ is retrieved, the search space could then be effectively reduced by assigning literals in the backbone to true.

In Definition 4, we shall introduce the definition of the biased weighted MAX-SAT instance, which will be used in the following proof.

**Definition 4.** Given a weighted MAX-SAT instance $wMAX-SAT(V,\tilde{C})$, the *biased weighted MAX-SAT* instance is defined as $wMAX-SAT(V,\bar{C})$, where $\bar{C} = \tilde{C} \cup \{x_i, \neg x_i | x_i \in V\}$ and the weight associated with every clause $\bar{C}_i \in \bar{C}$ is defined as follows.

(1) $\bar{w}(\bar{C}_i) = w(\bar{C}_i)$, if $\bar{C}_i \in \tilde{C}$ and $\bar{C}_i \notin \{x_j, \neg x_j | x_j \in V\}$;
(2) $\bar{w}(\bar{C}_i) = w(\bar{C}_i) + 1/2^{2j}$, if $\bar{C}_i = x_j \in \tilde{C}, x_j \in V$;
(3) $\bar{w}(\bar{C}_i) = w(\bar{C}_i) + 1/2^{2j+1}$, if $\bar{C}_i = \neg x_j \in \tilde{C}, x_j \in V$;
(4) $\bar{w}(\bar{C}_i) = 1/2^{2j}$, if $\bar{C}_i = x_j \notin \tilde{C}, x_j \in V$;
(5) $\bar{w}(\bar{C}_i) = 1/2^{2j+1}$, if $\bar{C}_i = \neg x_j \notin \tilde{C}, x_j \in V$.

Given a weighted MAX-SAT instance, it is easy to verify that the following property holds:
$\sum_{\bar{C}_i \in \bar{C}} \bar{w}(\bar{C}_i) = \sum_{\tilde{C}_i \in \tilde{C}} w(\tilde{C}_i) + 1/2^2 + 1/2^3 + \cdots + 1/2^{2n+1} = \sum_{\tilde{C}_i \in \tilde{C}} w(\tilde{C}_i) + 1/2 - 1/2^{2n+1}$.

## 3. Computational Complexity for Backbone

### 3.1 Weighted MAX-SAT and Biased Weighted MAX-SAT

In this section, we shall investigate the relationship between the solution of a weighted MAX-SAT instance and that of the biased weighted MAX-SAT instance in Lemma 1 and Lemma 2.

**Lemma 1.** Given the set of Boolean variables $V$ and the set of weighted clauses $\tilde{C}$, there exists a unique optimal solution to the biased weighted MAX-SAT instance $wMAX-SAT(V,\bar{C})$.

**Proof.** Given any two distinct solutions $s_1$, $s_2$ to the biased weighted MAX-SAT instance $wMAX-SAT(V,\bar{C})$, we will show that $w(V,\bar{C},s_1) \neq w(V,\bar{C},s_2)$.

Firstly, we construct a weighted MAX-SAT instance $wMAX-SAT(V,\{x_i, \neg x_i | x_i \in V\})$ where $w(x_i) = 1/2^{2i}, w(\neg x_i) = 1/2^{2i+1}$ for $x_i \in V$. We have that $w(V,\bar{C},s_1) = w(V,\tilde{C},s_1) + w(V,\{x_i, \neg x_i | x_i \in V\},s_1)$.

Since the weight $w(\tilde{C}_i)$ is a positive integer for each clause $\tilde{C}_i \in \tilde{C}$, thus $w(V,\tilde{C},s_1)$ must be the integer part of $w(V,\bar{C},s_1)$ and $w(V,\{x_i, \neg x_i | x_i \in V\},s_1)$ must be the fractional part of $w(V,\bar{C},s_1)$. Similarly, $w(V,\{x_i, \neg x_i | x_i \in V\},s_2)$ must be also the fractional part of $w(V,\bar{C},s_2)$. By assumption that $s_1 \neq s_2$, there must exist a variable $x_j \in V$ such that $x_j \in s_1 \cup s_2$ and $x_j \notin s_1 \cap s_2$. In the following proof, we only consider the case that $x_j \in s_1$, $\neg x_j \in s_2$. For the case that $\neg x_j \in s_1$ and $x_j \in s_2$, it can be proved in a similar way.

Obviously, the clause $x_j$ can be satisfied by $s_1$, while it cannot be satisfied by $s_2$. When viewed as binary encoded strings, the $2j$th bit of the fractional part of $w(V,\bar{C},s_1)$ will be 1, however the same bit of $w(V,\bar{C},s_2)$ will be 0. Hence, we have that $w(V,\bar{C},s_1) \neq w(V,\bar{C},s_2)$. Thus, this lemma is proved.

**Lemma 2.** Given the set of Boolean variables $V$ and the set of weighted clauses $\tilde{C}$, if $s^*$ is the unique optimal solution to the biased weighted MAX-SAT instance $wMAX-SAT(V,\bar{C})$, then $s^*$ is also optimal to weighted MAX-SAT instance $wMAX-SAT(V,\tilde{C})$.

**Proof.** Otherwise, there must exist a solution $s$ to the $wMAX-SAT(V,\tilde{C})$ such that $w(V,\tilde{C},s) > w(V,\tilde{C},s^*)$. By assumption that the weight $w(\tilde{C}_i)$ is a positive integer for each clause $\tilde{C}_i \in \tilde{C}$, thus we have $w(V,\tilde{C},s) \geq w(V,\tilde{C},s^*)+1$. Since $w(V,\bar{C},s) = w(V,\tilde{C},s) + w(V,\{x_i,\neg x_i | x_i \in V\},s)$, we have that $w(V,\bar{C},s) < w(V,\tilde{C},s)+1/2-1/2^{2n+1}$. Similarly, $w(V,\bar{C},s^*) < w(V,\tilde{C},s^*)+1/2-1/2^{2n+1}$. It implies that $w(V,\bar{C},s) > w(V,\tilde{C},s) \geq w(V,\tilde{C},s^*)+1 > w(V,\bar{C},s^*)$, which contradicts with the assumption that $s^*$ is the unique optimal solution to the $wMAX-SAT(V,\bar{C})$. Thus, this lemma is proved.

### 3.2 Computational Complexity for Retrieving Backbone

In Theorem 1, we shall show the intractability for retrieving the full backbone in weighted MAX-SAT. In addition, we shall present a stronger analytical result in Theorem 2 that it's NP-hard to retrieve a fixed fraction of the backbone.

**Theorem 1.** There exists no polynomial time algorithm to retrieve the backbone in weighted MAX-SAT unless $P = NP$.

**Proof.** Otherwise, there must exist an algorithm denoted by $\Lambda$ which is able to retrieve the backbone in weighted MAX-SAT in polynomial time.
Given any arbitrary weighted MAX-SAT instance $wMAX-SAT(V,\tilde{C})$, the biased weighted MAX-SAT instance $wMAX-SAT(V,\bar{C})$ can be constructed by an algorithm denoted by $\Gamma$ in $O(n)$ time.

Since the $wMAX-SAT(V,\bar{C})$ is also a weighted MAX-SAT instance, its backbone $bone(V,\bar{C})$ can be computed by $\Lambda$ in polynomial time (denoted by $O(\bullet)$). By Lemma 1, the backbone is the unique optimal solution to the $wMAX-SAT(V,\bar{C})$. By Lemm2, the $bone(V,\bar{C})$ is an optimal solution to the $wMAX-SAT(V,\tilde{C})$ as well.

Hence, any weighted MAX-SAT instance can be exactly solved in $O(n)+O(\bullet)$ time by $\Gamma$ and $\Lambda$. Obviously, such conclusion contradicts with the result that weighted MAX-SAT is NP-hard. Thus, this theorem is proved.

**Theorem 2.** There exists no polynomial time algorithm to retrieve a fixed fraction of the backbone in weighted MAX-SAT unless $P = NP$.

**Proof.** Otherwise, given any weighted MAX-SAT instance $wMAX-SAT(V,\tilde{C})$, we can always construct a biased weighted MAX-SAT instance $wMAX-SAT(V,\bar{C})$ according to Definition 4. As proven in Theorem 1, the backbone $bone(V,\bar{C})$ is an optimal solution to the $wMAX-SAT(V,\tilde{C})$.

If there exists a polynomial time algorithm $\Lambda$ to retrieve a fixed fraction of the backbone, we can acquire at least one literal $\ell$ in the $bone(V,\bar{C})$. In the following proof, we only consider the case that $\ell = x_j$ ($x_j \in V$). For the case that $\ell = \neg x_j$ ($x_j \in V$), it can be proved in a similar way.

Let $\hat{C}$ be the set of those clauses containing $\neg x_j$, let $\hat{C}'$ be the set of the remaining parts of the clauses from $\hat{C}$ after deleting $\neg x_j$. Since $\ell = x_j$, we can construct a new smaller weighted MAX-SAT instance $wMAX-SAT(V \setminus \{x_j\},\tilde{C}')$, where $\tilde{C}' = \{\tilde{C}_i | \tilde{C}_i \in \tilde{C}, \tilde{C}_i \text{ contains no } x_j \text{ or } \neg x_j\} \cup \hat{C}'$.

By repeating such procedures, an optimal solution to $wMAX-SAT(V,\tilde{C})$ can be finally obtained literal by literal in polynomial time, a contradiction. Hence, this theorem is proven.

## 4. Backbone Guided Local Search

### 4.1 Framework of Backbone Guided Local Search

In this section, we shall present the framework of backbone guided local search (BGLS) for weighted MAX-SAT. In BGSL, the local search process is guided by the backbone frequency, a group of probabilities representing the times of solution elements appearing in global optima. Since it's NP-hard to retrieve the exact backbone, we shall use the pseudo-backbone frequency instead, which can be computed by the approximate backbone (the common parts of local optima).

The framework of BGLS consists of two phases. The first phase (sampling phase) [13] collects the pseudo-backbone frequencies with traditional local search and the second one (backbone phase) obtains a solution with a backbone guided local search. Every run of local search is called a try. Each of two phases uses an input parameter to control the maximum tries of local search (see Algorithm 1).

---

**Algorithm 1**: BGLS for weighted MAX-SAT
**Input**: weighted MAX-SAT instance $wMAX-SAT(V,\tilde{C})$, $n_1$, $n_2$, a local search algorithm $H$
**Output**: solution $s^*$
**Begin**
  //sampling phase
  1. $w^* = -\infty$
  2. for $i=1$ to $n_1$ do
     2.1 obtain a solution $s_i$ with $H$;
     2.2 sample backbone from $s_i$;
     2.3 if $w(V,\tilde{C},s_i) > w^*$ then $w^* = w(V,\tilde{C},s_i), s^* = s_i$;
  //backbone phase
  3. for $i=1$ to $n_2$ do
     3.1 obtain a solution $s_i$ with backbone guided $H$;
     3.2 if $w(V,\tilde{C},s_i) > w^*$ then $w^* = w(V,\tilde{C},s_i), s^* = s_i$;
  4. return $s^*$;
**End**

---

Both of the two phases use the same local search algorithm. However, in the sampling phase, a particular step is used to sample backbone information from local optima. And the backbone phase uses this sampling to guide the following local search. Although many local search algorithms can be used as the algorithm $H$ in Algorithm 1, we proposed an improved Walksat in practice which is originally designed for MAX-SAT [15]. Thus, some necessary modifications should be conducted for adapting it to weighted MAX-SAT.

### 4.2 Walksat for Weighted MAX-SAT

In Algorithm 2, we shall present the Walksat for weighted MAX-SAT. In contrast to Walksat for MAX-SAT, some greedy strategies have been added to Algorithm 2 to fit weighted MAX-SAT. A try of Walksat needs two parameters. One is maximum number of flips that change the value of variables from true to false or inversely. The process of Walksat seeks repeatedly for a variable and flips it. When a variable is flipped, its break-count is defined as the number of satisfied clauses becoming unsatisfied. Obviously, the value of break-count will be nonnegative. And a variable with a zero break-count means the solution will be no worse than that before flipping. The other parameter is the probability for noise pick, which is used for determining when to use noise pick.

The Walksat for weighted MAX-SAT works as follows. After an initial assignment is generated by random values, Walksat picks a clause (clause pick) from one of the unsatisfied clauses with maximum weight randomly. Then the algorithm flips one of variables of this clause under the following rules: if there're more than one variable of zero break-count, then flip one of them randomly, otherwise flip any of variables randomly with the probability parameter (noise pick). For the case that the probability is not satisfied, pick one variable with least break-count randomly (greedy pick). Because the diversity of solutions depends on the parameter for noise pick and the static setting of it cannot react to the solving process [16-17]. So we choose a dynamic noise setting strategy (step 2.6) to set the parameter for noise pick: if the solution quality decreases, the noise parameter $p$ will increase to $p+(1-p)\cdot\varphi$, otherwise $p$ will decrease to $p-p\cdot\varphi/2$, where $\varphi$ is a predefined constant number.

In summary, there're five random actions (in step 1, 2.1, 2.3, respectively) during the whole process of the Walksat for weighted MAX-SAT. And we'll modify them in the backbone phase of BGLS (see Section 4.4).

---

**Algorithm 2**: Walksat for weighted MAX-SAT
**Input**: weighted MAX-SAT instance $wMAX-SAT(V,\tilde{C})$, $num$, a probability for noise pick $p$
**Output**: solution $s$
**Begin**
  1. let $s$ be a random assignment of variables of $V$, $w'=w(V,\tilde{C},s)$;
  2. for $i=1$ to $num-1$ do
    2.1 pick $c$, as one of the unsatisfied clauses with maximal weight, randomly;
    2.2 compute the break-count of all variables in $c$;
    2.3 if a zero break-count exists
        then pick one variable of zero break-count in $c$, randomly;
        else with a probability $p$, pick one of all variables in $c$, randomly;
            and with a probability $1-p$, pick one variable of least break-count ones in $c$, randomly;
    2.4 flip this variable and obtain solution $s'$;
    2.5 if $w(V,\tilde{C},s')>w'$ then $w'=w(V,\tilde{C},s'), s=s'$;
    2.6 adjust the value of $p$;
  3. return $s$;

---

### 4.3 Pseudo-backbone Frequencies Sampling

Given a particular problem instance, the exact difference between local optima and global optima cannot be exactly predicted unless those solutions are found, due to the diversity of instances. In this subsection, we analyze some typical instances to investigate the difference. The distance between local optima and global optima is defined as the number of different value of variables. For comparison among different instances, this distance is normalized by the total number of variables.

The sampling algorithm for local optima is Walksat for weighted MAX-SAT, where maximum of flips $num$ is set to be 200 and the noise probability $p$ is set to be 0. For every instance, the sampling algorithm was run for 50 times and 50 local optima were generated. Fig. 1 shows the distances between local optima and global optima for four instances [18]. The weight of solutions is plotted against the normalized distances to global optima. From Fig. 1, we can draw the conclusion that all the normalized distances between local optima and the global optimum are between 0.4 and 0.8. It implies that local optima and global optima have common values for most of variables. Thus, some approximate backbone can be retrieved to guide local search to find better solutions.

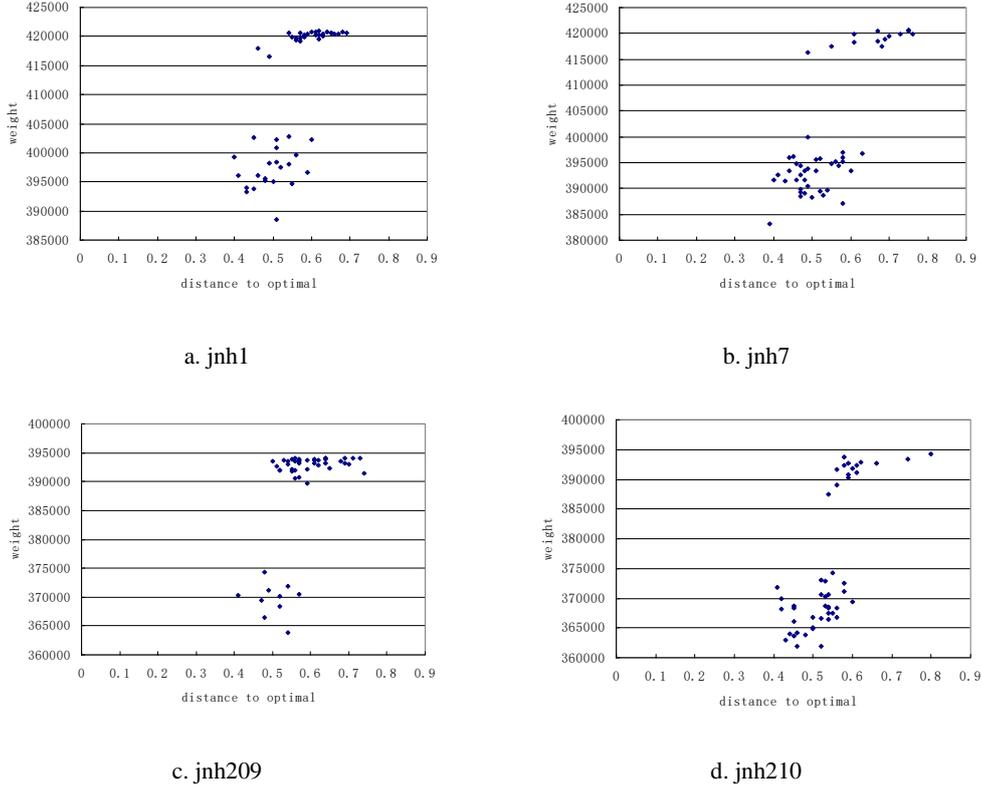

a. jnh1  b. jnh7

c. jnh209  d. jnh210

**Fig. 1 Relationship between local optima and global optima for 4 typical instances**

Furthermore, we recorded the number of every variable value in local optima and a probability of true or false being chosen for every variable can be computed. For every instance, we tested whether the variables in global optima tend to take the value with higher probability. And we found that more than 70% (74% for jnh1, 70% for jnh7, 73% for jnh209, and 73% for jnh210, respectively) of all variables in global optima take the values (true or false) with higher probabilities appearing in the local optima. It implies that a priority set of values can be made of those values with higher probability.

To guide the Walksat for weighted MAX-SAT, two forms of information should be recorded to construct the pseudo-backbone frequencies: the frequencies of variables and the frequencies of clauses. After obtaining any local optimum, the pseudo-backbone frequencies are updated. On one hand, the frequencies of variables record the times of the variable value appearing in local optima. On the other hand, the frequencies of clauses record the times of clauses being satisfied. The pseudo-backbone frequencies will be computed in the sampling phase (see Step 2.2 of Algorithm 1).

**4.4 Backbone Guided Walksat for Weighted MAX-SAT**

As discussed in Section 3.2, there're five random actions in the Walksat. Instead of the traditional random choosing, we shall determine the variable values by the pseudo-backbone frequencies (see Algorithm 3).

---

**Algorithm 3**: Backbone guided Walksat for weighted MAX-SAT
**Input**: weighted MAX-SAT instance $wMAX-SAT(V,\tilde{C})$, $num$, a probability for noise pick $p$, pseudo-backbone frequencies $F$
**Output**: solution $s$
**Begin**
1. let $s$ be a random assignment of variables of $V$, guided by $F$, $w' = w(V,\tilde{C},s)$;
2. for $i=1$ to $num-1$ do
   2.1 pick $c$ from $s$ as one of the unsatisfied clauses with maximum weight, guided by $F$;
   2.2 compute the break-count of all variables in $c$;
   2.3 if a zero break-count exists
       then pick one variable of zero break-count ones in $c$ randomly, guided by $F$;
       else with a probability $p$, pick one of all variables in $c$ randomly, guided by $F$;
           and with a probability $1-p$, pick one variable of least break-count ones in $c$, guided by $F$;
   2.4 flip this variable and obtain solution $s'$;
   2.5 if $w(V,\tilde{C},s') > w'$ then $w' = w(V,\tilde{C},s'), s = s'$;
   2.6 adjust the value of $p$;
3. return $s$;
**End**

---

The first random action is an initial assignment generation in Step 1 of Algorithm 3. An initial value of a variable can be guided by the frequencies of the values. The second random is in Step 2.1. In pseudo-backbone, the frequencies of clauses satisfied in local optima have been recorded. According to this, the probability of each clause being satisfied can be computed to direct which clause should be firstly satisfied. And then Step 2.3 includes three remaining random choices. These three choices are similarly guided by the same approximate backbone as the first random and the probabilities will be computed under the same method as the second random.

5. **Experimental Results**

In this section, we presented the experimental results for BGLS over weighted MAX-SAT benchmark. All the codes are implemented and compiled in C++ under a Pentium IV D 2.8GHz with 1GB memory. In the algorithm 1, 2 and 3, we have indicated 4 input parameters (the tries and maximum flips of two phases, respectively). In this experiment, we use identical parameters for both phases with 50 tries and 400 flips.

Since the best heuristics (GRASP/GRASP with path-relinking [8-9]) for weighted MAX-SAT are tested on distinct experimental platforms, we conducted a system performance conversion according to SPEC [19]. The system of GRASP with path-relinking is more than 15.69 times faster than ours (see Appendix) and only some of the instances' running time is available [9]. Since there's no platform of R4400 in SPEC, we give no running time of GRASP on PD2.8G.

The problem instances of weighted MAX-SAT [18] in the experiments include 44 instances, each of which has 100 variables and 800, 850 or 900 clauses, respectively. Tab. 1 shows the results of our experiments and the comparison with other algorithms. The column "Instance" indicates the input instances of benchmark and the column "Optima" shows the weight of global optima of instances. The following three columns "GRASP", "GRASP with path-relinking" and "BGLS" show the performance and results of the three algorithms, respectively. Each of them has sub-columns "weight" (the maximum weight obtained) and "time" (the time of running) for reporting the details. And the last column "Improvement" means the improvement of BGLS over GRASP, which is computed through dividing difference between the weight of BGLS and GRASP by the weight of global optima. A positive value means BGLS can obtain a better solution than GRASP, and vice verse.

From Tab.2, BGLS is more efficient than GRASP for nearly all the instances except instance "jnh208". The average of improvement is 0.0369%. For 19 instances out of 44, BGLS can always get global optima, however GRASP can get global optima for only 4 instances. And BGLS uses far less time than GRASP with path-relinking.

**Tab. 1 Experimental Results for Instances**

| Instance | Optima | GRASP | | | GRASP with path-relinking | | | BGLS | | Improvement |
|---|---|---|---|---|---|---|---|---|---|---|
| | weight | weight | time (seconds) | | weight | time (seconds) | | weight | time (seconds) | |
| | | | R4400 | PD2.8G | | Altix3700 | PD2.8G | | PD2.8G | |
| jnh1 | 420925 | 420737 | 192.1 | - | 420739 | 350 | 5491 | 420925 | 13 | 0.0447% |
| jnh4 | 420830 | 420615 | 467.9 | - | - | - | - | 420813 | 27 | 0.0470% |
| jnh5 | 420742 | 420488 | 30.9 | - | - | - | - | 420679 | 27 | 0.0454% |
| jnh6 | 420826 | 420816 | 504.2 | - | - | - | - | 420826 | 26 | 0.0024% |
| jnh7 | 420925 | 420925 | 188.1 | - | - | - | - | 420925 | 5 | 0.0000% |
| jnh8 | 420463 | 419885 | 546.4 | - | - | - | - | 420138 | 27 | 0.0602% |
| jnh9 | 420592 | 420078 | 41.2 | - | - | - | - | 420289 | 26 | 0.0502% |
| jnh10 | 420840 | 420565 | 591.1 | - | 420357 | 300 | 4707 | 420828 | 27 | 0.0625% |
| jnh11 | 420753 | 420642 | 757.4 | - | 420516 | - | - | 420672 | 26 | 0.0071% |
| jnh12 | 420925 | 420737 | 679.2 | - | 420871 | - | - | 420925 | 14 | 0.0447% |
| jnh13 | 420816 | 420533 | 12.9 | - | - | - | - | 420816 | 26 | 0.0673% |
| jnh14 | 420824 | 420510 | 197.7 | - | - | - | - | 420824 | 26 | 0.0746% |
| jnh15 | 420719 | 420360 | 424.6 | - | - | - | - | 420719 | 26 | 0.0853% |
| jnh16 | 420919 | 420851 | 392.8 | - | - | - | - | 420919 | 26 | 0.0162% |
| jnh17 | 420925 | 420807 | 448.0 | - | - | - | - | 420925 | 2 | 0.0280% |
| jnh18 | 420795 | 420372 | 142.9 | - | - | - | - | 420525 | 26 | 0.0364% |
| jnh19 | 420759 | 420323 | 611.3 | - | - | - | - | 420584 | 26 | 0.0620% |
| jnh201 | 394238 | 394238 | 604.3 | - | 394222 | 400 | 6276 | 394238 | 0 | 0.0000% |
| jnh202 | 394170 | 393983 | 348.7 | - | 393870 | - | - | 394029 | 25 | 0.0117% |
| jnh203 | 394199 | 393889 | 265.3 | - | - | - | - | 394135 | 25 | 0.0624% |
| jnh205 | 394238 | 394224 | 227.6 | - | - | - | - | 394238 | 4 | 0.0036% |
| jnh207 | 394238 | 394101 | 460.8 | - | - | - | - | 394238 | 4 | 0.0348% |
| jnh208 | 394159 | 393987 | 335.1 | - | - | - | - | 393819 | 25 | -0.0426% |
| jnh209 | 394238 | 394031 | 170.1 | - | - | - | - | 394238 | 7 | 0.0525% |
| jnh210 | 394238 | 394238 | 130.7 | - | - | - | - | 394238 | 4 | 0.0000% |
| jnh211 | 393979 | 393739 | 270.0 | - | - | - | - | 393979 | 24 | 0.0609% |

| | | | | | | | | | |
|---|---|---|---|---|---|---|---|---|---|
| jnh212 | 394238 | 394043 | 244.1 | - | 394006 | - | - | <u>394227</u> | 25 | 0.0467% |
| jnh214 | 394163 | 393701 | 486.4 | - | - | - | - | <u>394124</u> | 24 | 0.1073% |
| jnh215 | 394150 | 393858 | 601.8 | - | - | - | - | <u>394066</u> | 24 | 0.0528% |
| jnh216 | 394226 | 394029 | 441.1 | - | - | - | - | <u>394176</u> | 25 | 0.0373% |
| jnh217 | 394238 | 394232 | 125.4 | - | - | - | - | <u>394238</u> | 0 | 0.0015% |
| jnh218 | 394238 | 394099 | 155.7 | - | - | - | - | <u>394238</u> | 1 | 0.0353% |
| jnh219 | 394156 | 393720 | 502.8 | - | - | - | - | <u>393993</u> | 25 | 0.0693% |
| jnh220 | 394238 | 394053 | 513.5 | - | - | - | - | <u>394205</u> | 24 | 0.0386% |
| jnh301 | 444854 | 444670 | 522.1 | - | - | - | - | <u>444842</u> | 31 | 0.0387% |
| jnh302 | 444459 | 444248 | 493.9 | - | - | - | - | <u>444459</u> | 28 | 0.0475% |
| jnh303 | 444503 | 444244 | 416.1 | - | - | - | - | <u>444296</u> | 28 | 0.0117% |
| jnh304 | 444533 | 444214 | 96.7 | - | 444125 | 450 | 7060 | <u>444318</u> | 27 | 0.0234% |
| jnh305 | 444112 | 443503 | 424.9 | - | 443815 | 3500 | 54915 | <u>443533</u> | 29 | 0.0068% |
| jnh306 | 444838 | 444658 | 567.8 | - | 444692 | 2000 | 31380 | <u>444838</u> | 28 | 0.0405% |
| jnh307 | 444314 | 444159 | 353.5 | - | - | - | - | <u>444314</u> | 27 | 0.0349% |
| jnh308 | 444724 | 444222 | 50.2 | - | - | - | - | <u>444568</u> | 28 | 0.0778% |
| jnh309 | 444578 | 444349 | 86.8 | - | - | - | - | <u>444488</u> | 28 | 0.0313% |
| jnh310 | 444391 | 444282 | 44.7 | - | - | - | - | <u>444307</u> | 28 | 0.0056% |

## 6. Conclusions

In this paper, analytical results on the backbone in weighted MAX-SAT were presented ~~in this paper~~. We showed that it is intractable to retrieve the backbone in weighted MAX-SAT with any performance guarantee under the assumption that $P \neq NP$. And a backbone guided local search algorithm was proposed for weighted MAX-SAT.

Results of this paper imply a new way to incorporate the backbone in heuristics. The approximate backbone used to guide the flipping of literals in those local search based heuristics [9, 13, 14]. However, a multilevel strategy was introduced in the proof of Theorem 2. According to the proof, a new smaller weighted MAX-SAT instance could be constructed after a literal being fixed. By such a way, we can gradually reduce the original weighted MAX-SAT instance to a series of weighted MAX-SAT instances with less variables and clauses. Conversely, the solution to the original instance could be reconstructed with solutions of those reduced weighted MAX-SAT instances and fixed literals.

Many local search operators can apply on the BGLS for adapting with different problems. For weighted MAX-SAT, we used an improved Walksat to obtain local optima. Related to the common local search, BGLS heavily depends on the backbone sampled to simplify the scale of problems and to intensify local search. The process of retrieving backbone is to collect the feature from local optima. A note to design the similar algorithm is that a backbone guided algorithm can work well if and only if the local optima and the global optimum have the certain common parts.

In the future work, some interesting things remain to be investigated. Firstly, it needs further work on computational complexity for retrieving the backbone in the SAT. Although weighted MAX-SAT is a generalization of the SAT, it is not straightforward to prove the NP-hardness of retrieving the backbone in the SAT. The difficulty lies on the fact that the SAT isn't an optimization problem like weighted MAX-SAT but a typical combinatorial problem instead. For weighted MAX-SAT, we can always construct a weighted MAX-SAT instance with a unique optimal solution (i.e., the backbone) by slightly perturbing. However, such a method could not be directly applied to the SAT. Secondly, the ways for approximating the backbone are to be further explored. The backbone was approximated by the common parts of local optimal solutions [9, 13, 14]. However, we argue that there may exist better ways for approximating the backbone in weighted MAX-SAT. A good example of approximating the backbone by novel ways can be found in the traveling salesman problem (TSP) [20]. They proposed four strategies for approximating the backbone other than by the common parts of local optimal solutions: Minimum Spanning Tree, Nearest Neighbor, Lighter Than Median, and Close Pairs. By those new strategies, they claimed that very large TSP instances can be tackled with current state-of-the-art evolutionary local search heuristics. Inspired by their work, we shall approximate the backbone in weighted MAX-SAT in the future, by some similar methods which are generally exploited by exact algorithms.

**Appendix**

According to Tab. 2, the performance conversion from SPEC shows as follows: SGI Altix 3700: Intel 865P = 510/32.5 = 15.6923 > 15.69.

Tab. 2: Benchmark from SPEC

|  | SGI Altix 3700 Bx2 (1600 MHz 6M L3 Itaninum 2) | Intel 865P (2.8 GHz Pentium D) |
| --- | --- | --- |
| CINT 2000 Rates | 510 | 32.5 |

**References**

[1] Garey M R, Johnson D S. Computers and intractability: a guide to the theory of NP-completeness. San Francisco: W. H. Freeman, 1979. pp. 38

[2] Freuder E C, Wallace, R J. Partial constraint satisfaction. Artificial Intelligence, 1992, 58:21-70

[3] Hansen P, Jaumard B. Algorithms for the maximum satisfiability. Journal of Computing, 1990, 44:279-303

[4] Selman B, Levesque H and Mitchell D. A new method for solving hard satisfiability instances. In: Proceedings of the 10th National Conference on Artificial Intelligence. 1992, 440-446.

[5] Patrickmills, Edward T. Guided local search for solving SAT and weighted MAX-SAT problems. Journal of Automated Reasoning, 2000, 205-223

[6] Dalila B, Habiba D. Solving weighted MAX-SAT optimization problems using a taboo scatter search metaheuristic. In: Proceedings of the 2004 ACM symposium on Applied Computing, 2004.

[7] Habiba D, Aminc T, and Sofianc Z. Cooperative ant colonies for solving the maximum weighted satisfiability problem. IWANN 2003, Lecture Notes in Computer Science 2686, 2003, 446-453

[8] Resende M G C, Pitsoulis L S, Pardalos P M. Fortran subroutines for computing approximate solutions of weighted MAX-SAT problems using GRASP. Discrete Applied Mathematics, 2000, 100:95-113

[9] Festa P, Pardalos P M, Pitsoulis L S, Resende M G C. GRASP with path-relinking for the weighted maximum satisfiability problem. ACM Journal of Experimental Algorithms, 2006, 11:1-16

[10] Dubois O, Dequen G. A backbone-search heuristic for efficient solving of hard 3-SAT formula. In Proc. IJCAI-01, 2001, 248–253

[11] Zhang W X. Phase transition and backbone of the asymmetric traveling salesman problem. Journal of Artificial Intelligence Research, 2004, 21(1): 471-491.

[12] Jiang H, Zhang X C, Chen G L, and Li M C. Backbone analysis and algorithm design for the quadratic assignment problem. Science in China Series F-Information Science, 2008, 51(5):476 -488

[13] Telelis O, Stamatopoulos P. Heuristic backbone sampling for maximum satisfiability. In: Proceedings of the 2nd Hellenic Conference on Artificial Intelligence, 2002, 129-139

[14] Zhang W X, Rangan A, Looks M. Backbone guided local search for maximum satisfiability. In: Proceedings of the 18th International Joint Conference on Artificial Intelligence, 2003, 1179-1186

[15] Selman B, Kautz H, and Cohen B. Noise strategies for local search. In: Proc. AAAI-94.

[16] Hoos H H. An adaptive noise mechanism for Walksat. In: Proc. AAAI-02.

[17] Patterson D J, Kautz H. Auto-Walksat: A self-tuning implementation of Walksat. Electronic Notes in Discrete Mathematics, Elsevier, Amsterdam, vol.9, 2001, Presented at the LICS 2001 Workshop on Theory and Applications of Satisfiability Testing, June 14–15, Boston University, MA, 2001.

[18] Weighted MAX-SAT Problem Lib. http://www.research.att.com/~mgcr/data/maxsat.tar.gz.

[19] SPEC. http://www.spec.org/cgi-bin/osgresults?conf=cpu2000.

[20] Fischer T, Merz P. Reducing the size of traveling salesman problem instances by fixing edges. 7th European Conference on Evolutionary Computation in Combinatorial Optimization, Lecture Notes in Computer Science 4446, 2007, 72-83